\pdfoutput=1

\documentclass[11pt]{article}

\usepackage{acl}

\usepackage{times}
\usepackage{latexsym}

\usepackage[T1]{fontenc}

\usepackage[utf8]{inputenc}
\usepackage[main=english,russian]{babel}
\usepackage{booktabs}
\usepackage{graphicx}
\usepackage{colortbl}
\usepackage{enumitem}
\usepackage{caption}
\usepackage{adjustbox}
\usepackage{amssymb}
\usepackage{utfsym}
\usepackage{mathtools}
\usepackage{amsmath}
\usepackage{subcaption}

\usepackage{xcolor}

\newcommand{\greencheck}{{\color{green}\usym{2714}}}

\newcommand{\redcross}{{\color{red}\usym{2718}}}

\title{Exploring Methods for Cross-lingual Text Style Transfer: \\ The Case of Text Detoxification}

\author{
\textbf{Daryna Dementieva\textsuperscript{1}}\thanks{\hspace{3pt} Equal contribution}, \hspace{2pt} \textbf{Daniil Moskovskiy}\textsuperscript{2$\ast$}, \textbf{David Dale}\thanks{\hspace{3pt} Work has been done while at Skoltech} \hspace{2pt} \textbf{and}
\textbf{Alexander Panchenko\textsuperscript{2,3}} \\
\textsuperscript{1}Technical University of Munich,  
\textsuperscript{2}Skolkovo Institute of Science and Technology, \textsuperscript{3}AIRI\\
\href{mailto:daryna.dementieva@tum.de}{\small \textsf{ daryna.dementieva@tum.de}},
\href{mailto:a.panchenko@skol.tech}{\small \textsf{\{d.moskovskiy, a.panchenko\}@skol.tech}}
}

\begin{document}
\maketitle
\begin{abstract}

Text detoxification is the task of transferring the style of text from toxic to neutral. While there are approaches yielding promising results in monolingual setup, e.g., \cite{dale-etal-2021-text,DBLP:journals/corr/abs-2212-10543}, cross-lingual transfer for this task remains a challenging open problem \cite{moskovskiy-etal-2022-exploring}. In this work, we present a large-scale study of strategies for cross-lingual text detoxification -- given a parallel detoxification corpus for one language; the goal is to transfer detoxification ability to another language for which we do not have such a corpus.

Moreover, we are the first to explore a new task where text translation and detoxification are performed simultaneously, providing several strong baselines for this task. Finally, we introduce new automatic detoxification evaluation metrics with higher correlations with human judgments than previous benchmarks. We assess the most promising approaches also with manual markup, determining the answer for the best strategy to transfer the knowledge of text detoxification between languages. 
\end{abstract}

\section{Introduction}
The original monolingual task of text detoxification can be considered as text style transfer (TST), where the goal is to build a function that, given a source style $s^{src}$, a destination style $s^{dst}$, and an input text $t^{src}$ to produce an output text $t^{dst}$ such that: (i)~the style is indeed changed (in case of detoxification from \textsf{\small toxic} into \textsf{\small neutral}); (ii)~the content is saved as much as possible; (iii)~the newly generated text is fluent.


The task of detoxification was already addressed with several approaches. Firstly, several unsupervised methods based on masked language modelling  \cite{tran2020towards,dale-etal-2021-text} and disentangled representations for style and content \cite{DBLP:conf/acl/JohnMBV19,santos2018fighting} were explored. More recently, \citet{logacheva-etal-2022-paradetox} showed the superiority of supervised \textit{seq2seq} models for detoxification trained on a parallel corpus of crowdsourced {\small \textsf{toxic} $\leftrightarrow$ \textsf{neutral}} sentence pairs. Afterwards, there were experiments in multilingual detoxification. However, cross-lingual transfer between languages with multilingual \textit{seq2seq} models was shown to be a challenging task~\cite{moskovskiy-etal-2022-exploring}.

In this work, we aim to fill this gap and present an extensive overview of different approaches for cross-lingual text detoxification methods (tested in English and Russian), showing that promising results can be obtained in contrast to prior findings. Besides, we explore combining of two \textit{seq2seq} tasks/models in a single one to achieve computational gains (i.e., avoid the need to store and perform inference with several models). Namely, we conduct simultaneous translation and style transfer experiments, comparing them to a step-by-step pipeline. 

\begin{table}[h!]
    \centering
    \footnotesize
    \begin{tabular}{p{1.70cm}|p{5.3cm}}
    \toprule
        
        \multicolumn{2}{c}{\textbf{Monolingual Text Detoxification}} \\
        \midrule 
        Data & En parallel corpus \greencheck \\
        \midrule
        Original (En) \newline Detox (En) & Its a crock of s**t, and you know it. \newline It's quite unpleasant, and you know it. \\
        
        \midrule 
        
        \multicolumn{2}{c}{\textbf{Cross-lingual Detoxification Transfer (Ours \#1)}} \\
        \midrule
        Data  & En parallel corpus \greencheck, Ru parallel corpus \redcross \\
        
        \midrule
      
        Original (Ru) \newline Detox (Ru) & \foreignlanguage{russian}{Тварина е**ная, если это ее слова } \newline \foreignlanguage{russian}{Она очень неправа, если это действительно еще слова}\\
        
        \midrule
        
        \multicolumn{2}{c}{\textbf{Simultaneous  Detoxification\&Translation (Ours \#2)}} \\
        
        \midrule
        Data  &  
        En parallel corpus \greencheck, Ru parallel corpus \greencheck \\
      
        \midrule
        
        Original (Ru) \newline Detox (En) &  \foreignlanguage{russian}{Тварина е**ная, если это ее слова}  
        \newline She's not a good person if its her words \\
    \bottomrule
    \end{tabular}
    \caption{\textbf{Two new text detoxification setups} explored in this work compared to the monolingual setup.}
    \label{tab:intro_examples}
\end{table}

The contributions of this work are as follows:
\begin{itemize}
    \item We present a comprehensive study of \textit{cross-lingual detoxification transfer} methods,
    \item We are the first to explore the task of \textit{simultaneous detoxification and translation} and test several baseline approaches to solve it,
    \item We present a set of updated \textit{metrics for automatic evaluation} of detoxification improving correlations with human judgements.
\end{itemize}

\section{Related Work}

\paragraph{Text Detoxification Datasets} Previously, several datasets for different languages were released for toxic and hate speech detection. For instance, there exist several versions of Jigsaw datasets -- monolingual \cite{jigsaw_toxic} for English and multilingual \cite{jigsaw_multi} covering 6 languages. In addition, there are corpora specifically for Russian \cite{tox_ru_comments}, Korean \cite{DBLP:conf/acl-socialnlp/MoonCL20}, French \cite{DBLP:journals/information/VanetikM22} languages, \textit{inter alia}. These are non-parallel classification datasets. In previous work on detoxification methods, such kind of datasets were used to develop and test unsupervised text style transfer approaches \cite{maskandinfill, tran2020towards, dale-etal-2021-text, DBLP:journals/corr/abs-2212-10543}.

However, lately a parallel dataset \textit{ParaDetox} for training supervised text detoxification models for English was released \cite{logacheva-etal-2022-paradetox} similar to previous parallel TST datasets for formality~\cite{rao-tetreault-2018-dear,briakou-etal-2021-ola}. Pairs of toxic-neutral sentences were collected with a pipeline based on three crowdsourcing tasks. The first task is the main paraphrasing task. Then, the next two tasks -- content preservation check and toxicity classification -- are used to verify a paraphrase. 
Using this crowdsourcing methodology, a Russian parallel text detoxification dataset was also collected \cite{russe2022detoxification}. We base our cross-lingual text detoxification experiments on these comparably collected data (cf. Table~\ref{tab:datasets}).

\paragraph{Text Detoxification Models} Addressing text detoxification task as \textit{seq2seq} task based on a parallel corpus was shown to be more successful than the application of unsupervised methods by \citet{logacheva-etal-2022-paradetox}. For English methods, the fine-tuned BART model \cite{lewis-etal-2020-bart} on English ParaDetox significantly outperformed all the baselines and other \textit{seq2seq} models in both automatic and manual evaluations. For Russian  in \cite{russe2022detoxification}, there was released ruT5 model \cite{DBLP:journals/jmlr/RaffelSRLNMZLL20} fined-tuned on Russian ParaDetox. 
These SOTA monolingual models for English\footnote{\href{https://huggingface.co/s-nlp/bart-base-detox}{https://huggingface.co/s-nlp/bart-base-detox}} and Russian\footnote{\href{https://huggingface.co/s-nlp/ruT5-base-detox}{https://huggingface.co/s-nlp/ruT5-base-detox}} are publicly available.

\paragraph{Multilingual Models}


Together with pre-trained monolingual language models (LM), there is a trend of releasing multilingual models covering more and more languages. For instance, the NLLB model \cite{costa2022no} is pre-trained for 200 languages. However, large multilingual models can have many parameters (NLLB has 54.5B parameters), simultaneously requiring a vast amount of GPU memory to work with it. 

As the SOTA detoxification models were fine-tuned versions of T5 and BART, we experiment in this work with multilingual versions of them -- \textbf{mT5} \cite{xue2021mt5} and \textbf{mBART} \cite{tang2020multilingual}. The mT5 model covers 101 languages and has several versions.
The mBART model has several implementations and several versions as well. We use mBART-50, which covers 50 languages. Also, we use in our experiments the \textbf{M2M100} model \cite{DBLP:journals/jmlr/FanBSMEGBCWCGBL21} that was trained for translation between 100 languages. All these models have less than 1B parameters (in \textit{large} versions). 

\paragraph{Cross-lingual Knowledge Transfer} 
A common case is when data for a specific task is available for English but none for the target language. In this situation, techniques for knowledge transfer between languages are applied.

\begin{table}[t!]
    \centering
    \footnotesize
    \resizebox{0.49\textwidth}{!}{
        \begin{tabular}{p{4.3cm}|c|c|c|c}
        \toprule
             & Train & Dev & Test & Total \\
            \midrule
            English \cite{logacheva-etal-2022-paradetox} & $18\,777$ & $988$ & $671$ & $20\,436$ \\
            \hline
            Russian \cite{russe2022detoxification} & $5\,058$ & $1\,000$ & $1\,000$ & $7\,058$ \\
        \bottomrule
        \end{tabular}}
    \caption{\textbf{Parallel datasets for text detoxification} used in our cross-lingual detoxification experiments.}
    \label{tab:datasets}
\end{table}

One of the approaches usually used to address the lack of training data is the translation approach. It was already tested for offensive language classification \cite{el2022multilingual,DBLP:journals/csse/WadudMSNS23}. The idea is to translate the training data in the available language into the target language and train the corresponding model based on the new translated dataset.

\begin{figure*}[th!]
    \centering
    \includegraphics[width=0.9\textwidth]{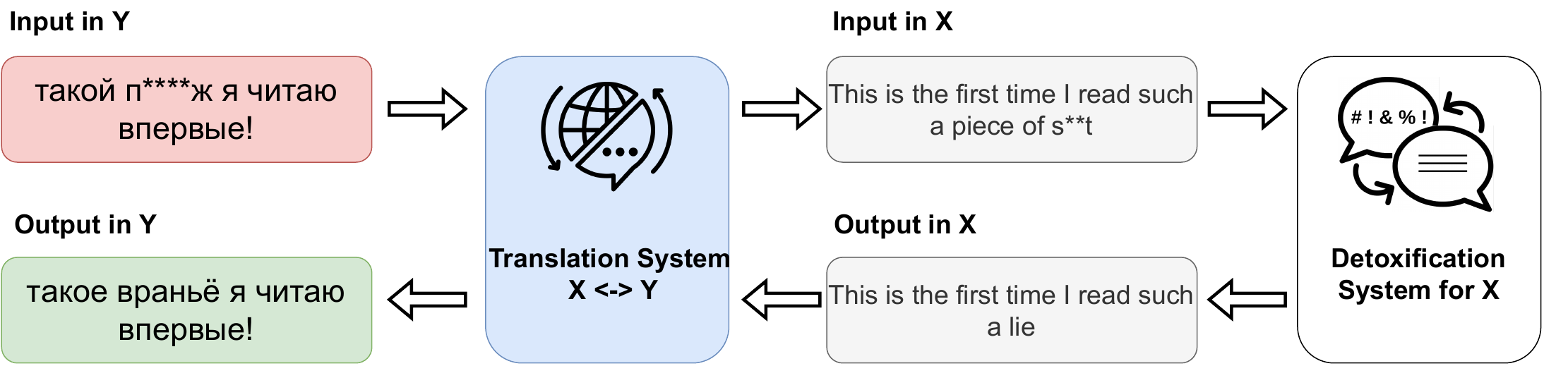}
    \caption{\textbf{Backtranslation approach}: (i) translate input text into resource-rich language; (ii) perform detoxification; (iii) translate back into target language.}
    \label{fig:backtranslation}
\end{figure*}

The methods for zero-shot and few-shot text style transfer were already explored. In \cite{DBLP:conf/acl/KrishnaNGST22}, the operation between style and language embeddings is used to transfer style knowledge to a new language. The authors in \cite{DBLP:conf/acl/LaiTN22} use adapter layers to incorporate the knowledge about the target language into a TST model.

For text detoxification, only in \cite{moskovskiy-etal-2022-exploring} cross-lingual setup was explored through the translation of inputs and outputs of a monolingual system. It has been shown that detoxification trained for English using a multilingual Transformer is not working for Russian (and vice versa). In this work, we present several approaches to cross-lingual detoxification, which, in contrast, yield promising results.

\paragraph{Simultaneous Text Generation\&Translation} The simultaneous translation and text generation was already introduced for text summarization. Several datasets with a wide variety of languages were created \cite{DBLP:conf/emnlp/Perez-Beltrachini21,DBLP:journals/corr/abs-2112-08804}. The main approaches to tackle this task -- either to perform step-by-step text generation and translation or train a supervised model on a parallel corpus. To the best of our knowledge, there were no such experiments in the domain of text detoxification. This work provides the first experiments to address this gap.

\section{Cross-lingual Detoxification Transfer}
In this section, we consider the setup when a parallel detoxification corpus is available for a resource-rich language (e.g., English), but we need to perform detoxification for another language such corpus is unavailable. We test several approaches that differ by the amount of data and computational sources listed below.

\subsection{Backtranslation}
One of the baseline approaches is translating input sentences into the language for which a detoxification model is available. For instance, we first train a detoxification model on available English ParaDetox. Then, if we have an input sentence in another language, we translate it into English, perform detoxification, and translate it back into Russian (Figure~\ref{fig:backtranslation}). Thus, for this approach, we require two models (one model for translation and one for detoxification) and three inferences (one for translation from the target language into the available language, text detoxification, and translation back into the target language).

In previous work \cite{moskovskiy-etal-2022-exploring}, \textbf{Google Translate API} and \textbf{FSMT} \cite{DBLP:conf/wmt/NgYBOAE19} models were used to make translations. In this work, we extend these experiments with two additional models for translation: 
\begin{itemize}
    \item \textbf{Helsinki OPUS-MT} \cite{DBLP:conf/eamt/TiedemannT20} -- Transformer-based model trained specifically for English-Russian translation.\footnote{\href{https://huggingface.co/Helsinki-NLP/opus-mt-ru-en}{https://huggingface.co/Helsinki-NLP/opus-mt-ru-en}}
    \item \textbf{Yandex} Translate API available from Yandex company and considered high/top quality for the Russian-English pair.\footnote{\href{https://tech.yandex.com/translate}{https://tech.yandex.com/translate}}
\end{itemize}

We test the backtranslation approach with two types of models: (i)~SOTA models for corresponding monolingual detoxification; (ii)~multilingual LM.

\subsection{Training Data Translation}
\label{sec:training_data_translation}

\begin{figure}[th!]
    \centering
    \includegraphics[width=0.45\textwidth]{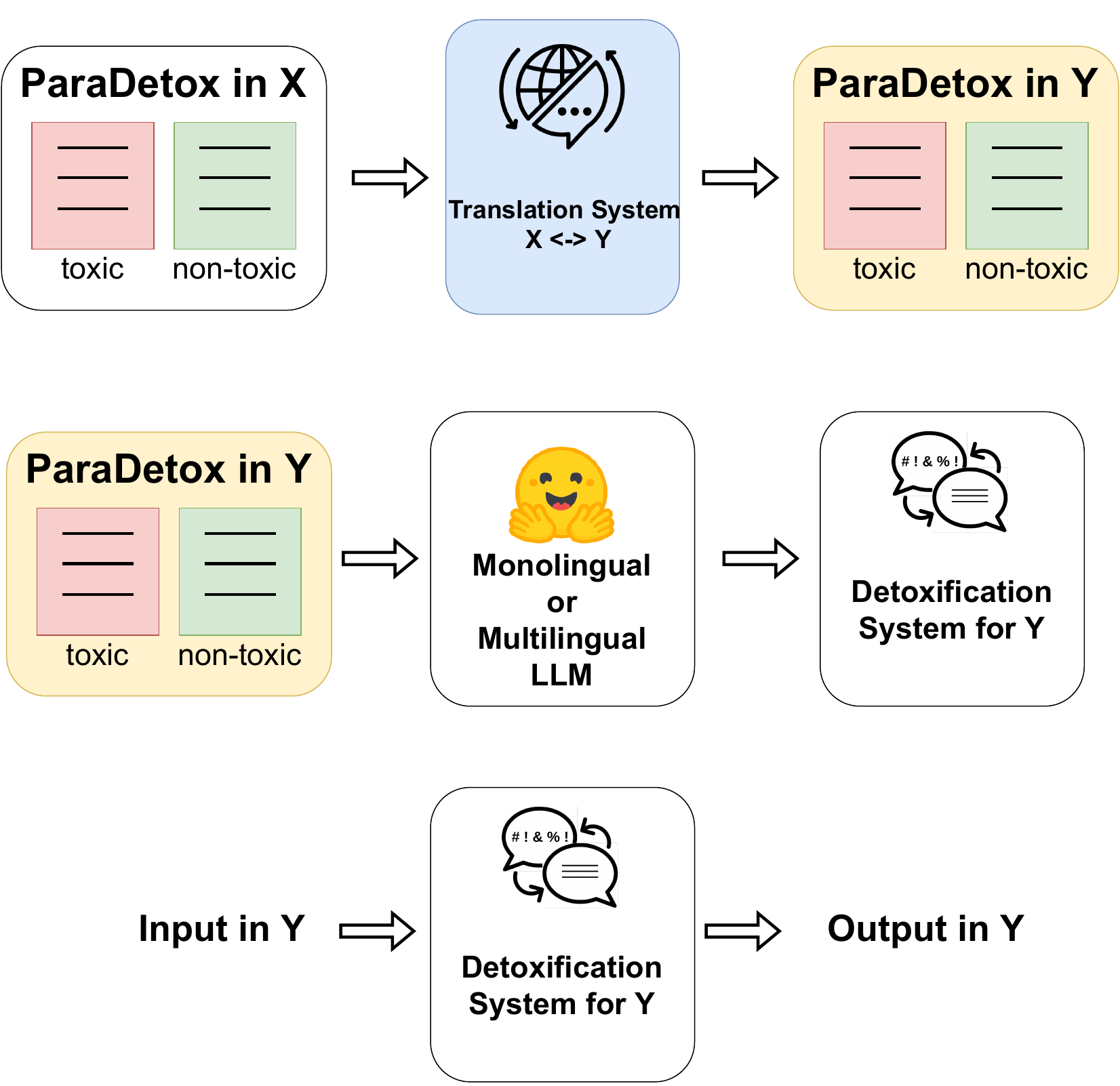}
    \caption{\textbf{Training Data Translation approach}: (i) translate available dataset into the target language; (ii) train detoxification model for the target language.}
    \label{fig:corpus_translation}
\end{figure}

Another way of how translation can be used is the translation of available training data. If we have available training data in one language, we can fully translate it into another and use it to train a separate detoxification model for this language (Figure~\ref{fig:corpus_translation}). For translation, we use the same models described in the previous section.


As detoxification corpus is available for the target language in this setup, we can fine-tune either multilingual LM where this language is present or monolingual LM if it is separately pre-trained for the required language. Compared to the previous approach, this method requires a fine-tuning step that implies additional computational resources.


\subsection{Multitask Learning}
Extending the idea of using translated ParaDetox, we can add additional datasets that might help improve model performance. 

We suggest multitasking training for cross-lingual detoxification transfer. We take a multilingual LM where resource-rich and target languages are available. Then, for the training, we perform multitask procedure which is based on the following tasks: (i)~translation between the resource-rich language and target language; (ii)~paraphrasing for the target language; (iii) detoxification for the resource-rich language for which original ParaDetox is available; (iv) detoxification for the target language based on translated data.

Even if the LM is already multilingual, we suggest that the translation task data help strengthen the bond between languages. As the detoxification task can be seen as a paraphrasing task as well, the paraphrasing data for the target language can add knowledge to the model of how paraphrasing works for this language. Then, the model is basically trained for the detoxification task on the available data.

\subsection{Adapter Training}
\begin{figure}[th!]
    \centering
    \includegraphics[width=0.45\textwidth]{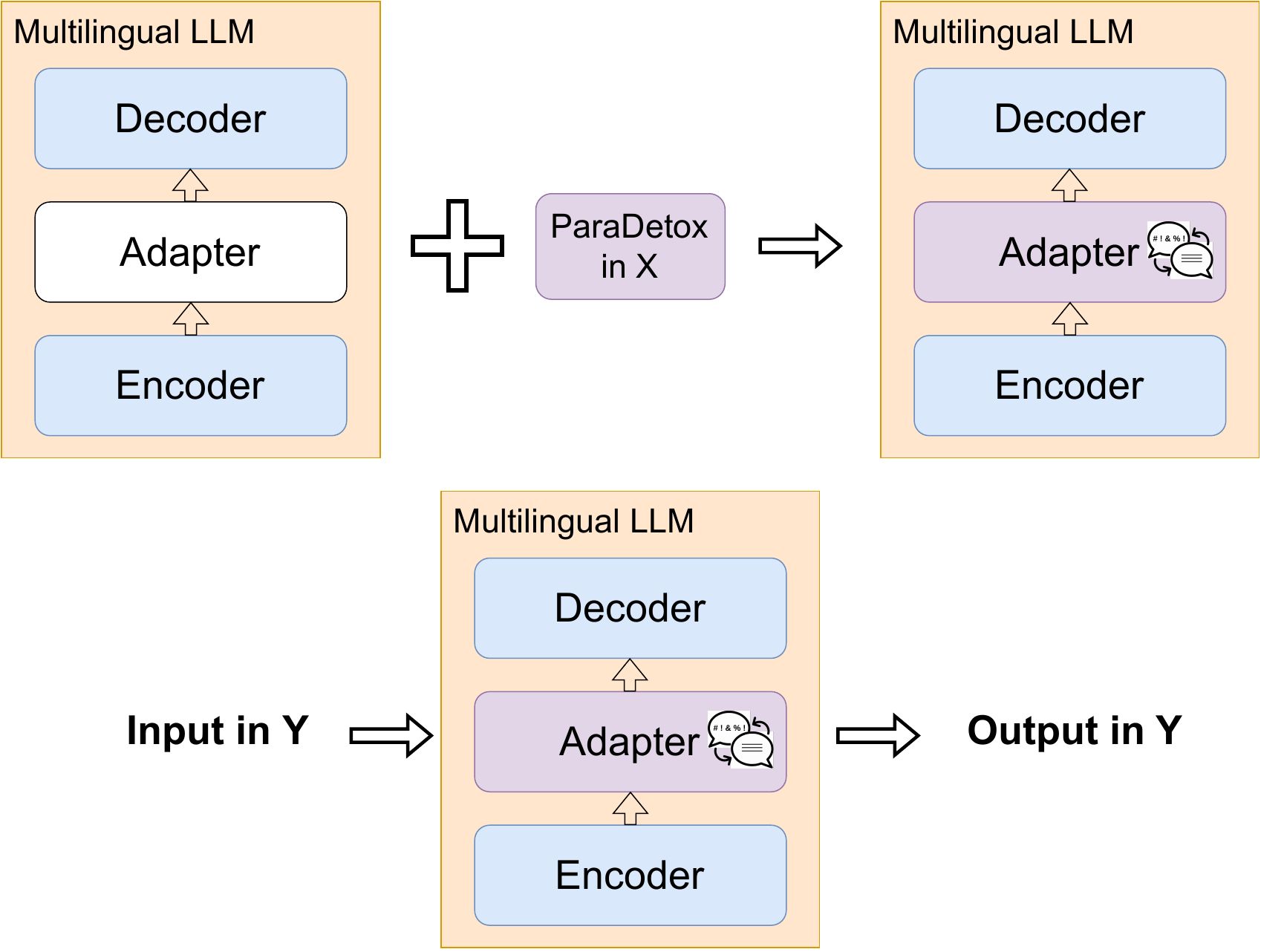}
    \caption{\textbf{Adapter approach}: (i) insert Adapter layer into Multilingual LM; (ii) train only Adapter for detoxification task on the available corpus.}
    \label{fig:adapter}
\end{figure}

For paraphrasing corpus, we use \textbf{Opusparcus} corpus \cite{DBLP:conf/lrec/Creutz18}. For translation, we use corresponding \texttt{en-ru} parts of \textbf{Open Subtitles} \cite{DBLP:conf/lrec/LisonT16}, \textbf{Tatoeba} \cite{DBLP:conf/wmt/Tiedemann20}, and \textbf{news\_commentary}\footnote{\href{https://huggingface.co/datasets/news_commentary}{https://huggingface.co/datasets/news\_commentary}} corpora.

To eliminate the translation step, we present a new approach based on the Adapter Layer idea \cite{DBLP:conf/icml/HoulsbyGJMLGAG19}. The usual pipeline of \textit{seq2seq} generation process is:
\begin{align}
    y = \text{Decoder}(\text{Encoder}(x))
\end{align}

We add an additional Adapter layer in the model:
\begin{align}
    y = \text{Decoder}(\text{Adapter}(\text{Encoder}(x))),
\end{align}
where $Adapter = Linear(ReLU(Linear(x)))$ and gets as input the output embeddings from encoder. 

Any multilingual pre-trained model can be taken for a base \textit{seq2seq} model. Then, we integrate the Adapter layer between the encoder and decoder blocks. For the training procedure, we train the model on a monolingual ParaDetox corpus available. However, we do not update all the weights of all model blocks, only the Adapter. As a result, we force the Adapter layer to learn the information about detoxification while the rest of the blocks save the knowledge about multiple languages. We can now input the text in the target language during inference and obtain the corresponding detoxified output (Figure~\ref{fig:adapter}). Compared to previous approaches, the Adapter training requires only one model fine-tuning procedure and one inference step. While in \cite{DBLP:conf/acl/LaiTN22} there were used several Adapter layers pre-trained specifically for the language, we propose to use only one layer between the encoder and decoder of multilingual LM that will incorporate the knowledge about the task.

For this approach, we experiment with the \textbf{M2M100} and \textbf{mBART-50} models. While the M2M100 model is already trained for the translation task, this version of mBART is pre-trained only on the denoising task. Thus, we additionally pre-train this model on paraphrasing and translation corpora used for the Multitask approach. During the training and inference with the mBART model, we explicitly identify which language the input and output are given or expected with special tokens.



\section{Detox\&Translation}
The setup of simultaneous detoxification and translation occurs when the toxic and non-toxic parts of the training parallel dataset are in different languages. For instance, a toxic sentence in a pair is in English, while its non-toxic paraphrase is in Russian.

\begin{figure}[bp!]
    \centering
    \includegraphics[width=0.5\textwidth]{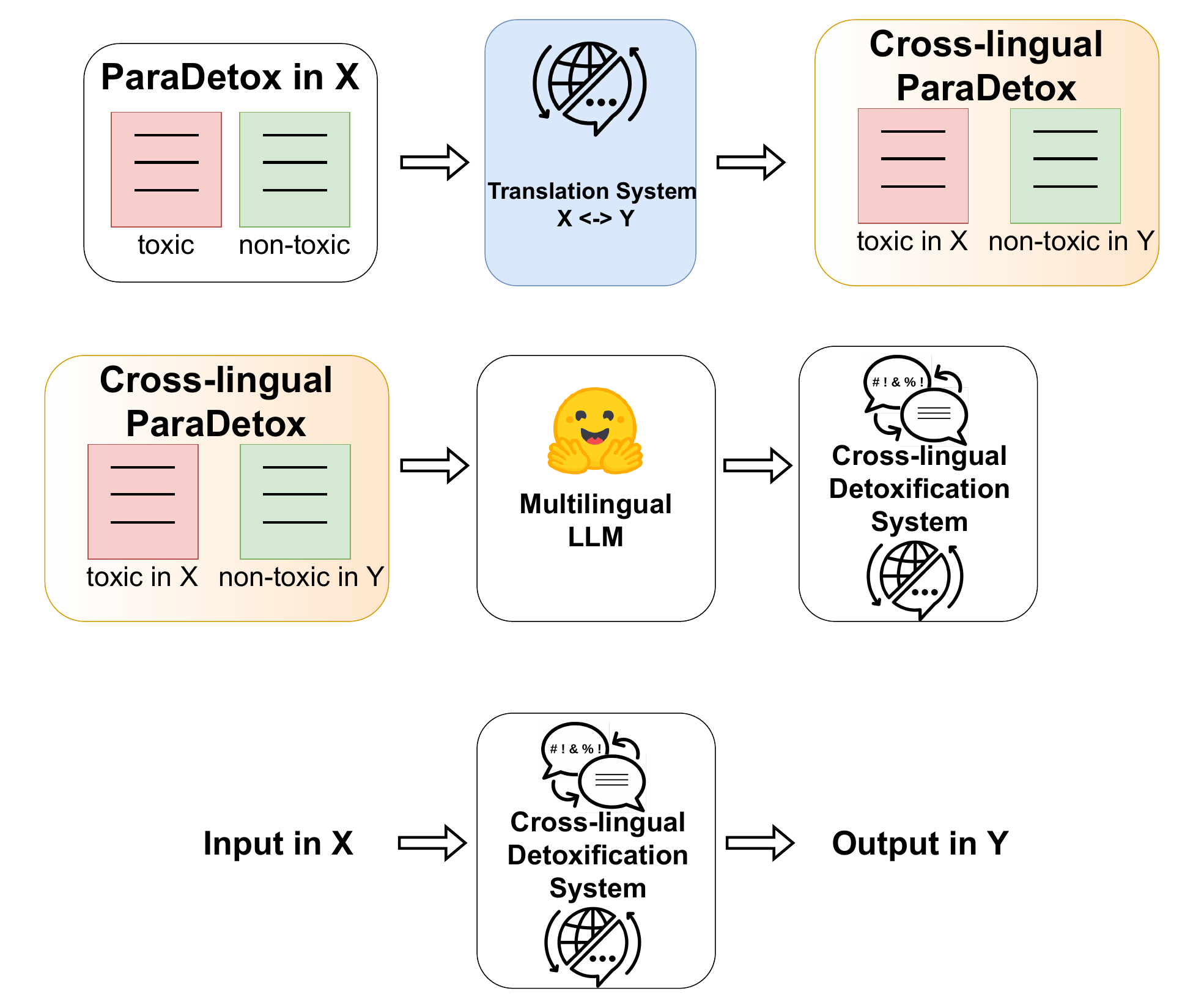}
    \caption{\textbf{Simultaneous Detox\&Translate} approach is based on synthetic cross-lingual parallel corpus.}
    \label{fig:cross_lingual_inference}
\end{figure}

The baseline approach to address text detoxification from one language to another can be to perform step-by-step detoxification and translation. However, that will be two inference procedures, each potentially with a computationally heavy seq2seq model. To save resources for one inference, in this section, we explore the models that can perform detoxification and translation in one step.

While for cross-lingual text summarization, parallel datasets were obtained, there are no such data for text detoxification. The proposed approach is creating a synthetic cross-lingual detoxification dataset (Figure~\ref{fig:cross_lingual_inference}). Then, we train simultaneously model for detoxification as well as for translation. The models described in the section above were also used for the translation step of parallel corpora.

\section{Evaluation Setups}
There are plenty of work developing systems for text detoxification. Yet, in each work, the comparison between models is made by automatic metrics that are not unified, and their choice may be arbitrary \cite{DBLP:journals/corr/abs-2306-00539}. There are several recent works that studied the correlation between automatic and manual evaluation for text style transfer tasks -- formality \cite{lai-etal-2022-human} and toxicity \cite{logacheva-etal-2022-study}. Our work presents a new set of metrics for automatic evaluation for English and Russian languages, confirming our choice with correlations with manual metrics.

For all languages, the automatic evaluation consists of three main parameters:
\begin{itemize}
    \item \textit{Style transfer accuracy} (\textbf{STA$_a$}): percentage of non-toxic outputs identified by a style classifier. In our case, we train for each language corresponding toxicity classifier.
    \item \textit{Content preservation} (\textbf{SIM$_a$}): measurement of the extent to which the content of the original text is preserved.
    \item \textit{Fluency} (\textbf{FL$_a$}): percentage of fluent sentences in the output.
\end{itemize}

The aforementioned metrics must be properly combined to get one \textit{Joint} metric to rank models. We calculate \textbf{J} as following:
\begin{equation}
    \textbf{J} = \frac{1}{n}\sum\limits_{i=1}^{n}\textbf{STA}(x_i) \cdot \textbf{SIM}(x_i) \cdot \textbf{FL}(x_i), 
\end{equation}

where the scores \textbf{STA}($x_i$), \textbf{SIM}($x_i$), \textbf{FL}($x_i$) $\in \{0, 1\}$ meaning the belonging to the corresponding class. 


\subsection{Automatic Evaluation for English}
Our setup is mostly based on metrics previously used by \cite{logacheva-etal-2022-paradetox}: only the content similarity metric is updated as other metrics obtain high correlations with human judgments.  

\paragraph{Style accuracy} STA$_a$ metric is calculated with a  RoBERTa-based \cite{roberta}  style classifier trained on the union of three Jigsaw datasets \cite{jigsaw_toxic}.

\paragraph{Content similarity} Before, SIM$_a^{old}$ was estimated as cosine similarity between the embeddings of the original text and the output computed with the model of~\citep{wieting-etal-2019-beyond}. This model is trained on paraphrase pairs extracted from ParaNMT~\citep{wieting-gimpel-2018-paranmt} corpus.

We propose to estimate SIM$_a$ as BLEURT score \cite{sellam-etal-2020-bleurt}.
In \cite{babakov-etal-2022-large}, a large investigation on similarity metrics for paraphrasing and style transfer tasks. The results showed that the BLEURT metric has the highest correlations with human assessments for text style transfer tasks for the English language.

\paragraph{Fluency} FL$_a$ is the percentage of fluent sentences identified by a RoBERTa-based classifier of linguistic acceptability trained on the CoLA dataset~\citep{cola}.

\subsection{Automatic Evaluation for Russian}

The set of previous and our proposed metrics is listed below (the setup to compare with is based on \cite{russe2022detoxification}):

\paragraph{Style accuracy} In \cite{russe2022detoxification}, STA$_a^{old}$ is computed with a RuBERT Conversational classifier \cite{kuratov2019adaptation} fine-tuned on Russian Language Toxic Comments dataset collected from \url{2ch.hk} and Toxic Russian Comments dataset collected from \url{ok.ru}.

In our updated metric STA$_a$, we change the toxicity classifier using the more robust to adversarial attacks version presented in \cite{DBLP:journals/corr/abs-2204-13638}.

\paragraph{Content similarity} Previous implementation of SIM$_a^{old}$ is evaluated as a cosine similarity of LaBSE \citep{feng2020languageagnostic} sentence embeddings. 

The updated metric SIM$_a$ is computed as a classifier score of RuBERT Conversational fine-tuned for paraphrase classification on three datasets:
Russian Paraphrase Corpus \cite{gudkov-etal-2020-automatically}, RuPAWS \cite{DBLP:conf/lrec/MartynovKLPKS22}, and content evaluation part from Russian parallel corpus \cite{russe2022detoxification}.

\paragraph{Fluency} Previous metric FL$_a^{old}$ is measured with a BERT-based classifier \citep{devlin-etal-2019-bert} trained to distinguish real texts from corrupted ones. The model was trained on Russian texts and their corrupted (random word replacement, word deletion, insertion, word shuffling, etc.) versions.

In our updated metric FL$_a$, to make it symmetric with the English setup, fluency for the Russian language is also evaluated as a RoBERTa-based classifier fine-tuned on the language acceptability dataset for the Russian language RuCoLA \cite{DBLP:journals/corr/abs-2210-12814}.

\begin{table}[t!]
    \centering
    \small
    \begin{tabular}{c|c|c}
    \toprule
    & \textbf{Old metrics} & \textbf{Ours metrics} \\
        \midrule
        STA & 0.472 & \textbf{0.598} \\
        SIM & 0.124 & \textbf{0.244} \\
        FL  & -0.011 & \textbf{0.354} \\
        J   & 0.106 & \textbf{0.482} \\
    \bottomrule
    \end{tabular}
    \caption{\textbf{Ours vs old evaluation setups}. Spearman's correlation between automatic vs manual setups for each old and new evaluation parameter based on systems scores for \textit{Russian} language. All numbers denote the statistically significant correlation ($p$-value $\leq 0.05$).}
    \label{tab:manual_vs_auto_ru_short}
\end{table}

\begin{table*}[ht!]
    \centering
    \footnotesize
    \begin{tabular}{p{2cm}|p{4cm}|p{4cm}|c|c|c}
    \toprule
        \textbf{Method} & \textbf{Models} & \textbf{Datasets} & \textbf{\shortstack{Data \\ Creation}} & \textbf{\shortstack{Fine \\ tuning}} & \textbf{\shortstack{\# Inference \\ Steps }} \\
        \hline
        \textit{Backtranslation} & - Detoxification model for the resource-rich language; \newline - Translation model to the target language; & \multicolumn{1}{c|}{---} & \redcross & \redcross & 3 \\
        \hline
        \textit{Training Data Translation} & - Translation model to the target language; \newline - Auto-regressive multilingual or monolingual LM for the target language; & - ParaDetox on the resource-rich language; & \greencheck & \greencheck & 1 \\
        \hline
        \textit{Multitask Learning} & - Auto-regressive multilingual or monolingual LM for the target language; & - ParaDetox on the resource-rich language; \newline - Corpus for translation between the resource-rich and target languages; \newline - Corpus for paraphrasing for the target language; & \greencheck & \greencheck & 1 \\
        \hline
        \textit{Adapter \newline Training} & - Auto-regressive multilingual LM where the resource-rich and target languages are present; & - ParaDetox on the resource-rich language; \newline - Corpus for translation between the resource-rich and target languages; \newline - Corpus for paraphrasing for the target language; & \redcross & \greencheck & 1 \\
    \bottomrule
    \end{tabular}
    \caption{Comparison of the proposed approaches for cross-lingual detoxification transfer based on required computational and data resources. As one may observe, backtranslation approach requires 3 runs of seq2seq models, while other approaches are based on a single (end2end) model and require only one run. }
    \label{tab:resources_approaches}
\end{table*}

We use the manual assessments available from \cite{russe2022detoxification} to calculate correlations with manual assessments. We have 850 toxic samples in the test set evaluated manually via crowdsourcing by three parameters -- toxicity, content, and fluency.  
%
We can see in Table~\ref{tab:manual_vs_auto_ru_short} the correlations between human assessments and new metrics are higher than for the previous evaluation setup (see details in Appendix~\ref{sec:app_correlations}). 


To calculate \textbf{SIM} metric for \textbf{Detox\&Translation} task
we use the monolingual version of SIM for the target language, comparing the output with the input translated into the target language. For instance, if Detox\&Translation is done from English to Russian, we translate English toxic input to Russian language and compare it with the output using Russian SIM$_a$.

\subsection{Manual Evaluation}
As the correlation between automatic and manual scores still has room for improvement, we also evaluate selected models manually. We invited three annotators fluent in both languages to markup the corresponding three parameters of evaluation (instructions in Appendix~\ref{sec:app_manual_instructions}). A subset of 50 samples from the corresponding test sets were randomly chosen for this evaluation. The interannotator agreement (Krippendorff’s $\alpha$) reaches 0.74 (STA), 0.60 (SIM), and 0.71 (FL).

\section{Results}

The \textbf{automatic evaluation} results are presented in Table~\ref{tab:cross_lingual_results}. Together with the metrics evaluation, we also assess the proposed methods based on the required resources (Table~\ref{tab:resources_approaches}). We take test sets provided for both English and Russian datasets for evaluation (as presented in Table~\ref{tab:datasets}). Firstly, we report scores of humans reference and trivial duplication of the input toxic text. Then, we present strong baselines based on local edits -- Delete and condBERT \cite{dale-etal-2021-text, DBLP:journals/mti/DementievaMLDKS21} -- and, finally, SOTA \textit{seq2seq} detoxification monolingual models  based on T5/BART. Moreover, we report the performance of multilingual models (mBART/M2M100) trained on monolingual parallel corpus separately (RU/EN) or on the joint corpus (RU+EN) to check the credibility of training multilingual models for such a task. The results of the \textbf{manual evaluation} are reported in Table~\ref{tab:manual_results} comparing only the best models identified with automatic evaluation. 

Additional results are available in appendices: Appendix~\ref{sec:app_examples} contains examples of models' outputs; Appendix~\ref{sec:app_translation_examples} contains examples of toxic text translations; Appendix~\ref{sec:app_translation} presents a comparison of different translation methods for each approach.


\begin{table*}[ht!]
\footnotesize
\centering

\begin{tabular}{p{3.65cm}|c|c|c|c|c|c|c|c}
\toprule

& \textbf{STA} & \textbf{SIM} & \textbf{FL} & \textbf{J} & \textbf{STA} & \textbf{SIM} & \textbf{FL} & \textbf{J} \\ \hline
& \multicolumn{4}{c|}{{Russian}} & \multicolumn{4}{c}{{English}} \\ \hline \hline
\multicolumn{9}{c}{\textbf{Baselines: Monolingual Setup} (on a language with a parallel corpus)} \\ \hline
Human references & 0.788 & 0.733 & 0.820 & 0.470  & 0.950 & 0.561 & 0.836 & 0.450  \\
Duplicate input & 0.072 & 0.785 & 0.783 & 0.045 & 0.023 & 0.726 & 0.871 & 0.015  \\ \hline
\multicolumn{9}{c}{\it Monolingual models trained on monolingual parallel corpus} \\ \hline
Delete & 0.408 & 0.761 & 0.700 & 0.210  & 0.815 & 0.574 & 0.690 & 0.308  \\
condBERT & 0.654 & 0.671 & 0.579 & 0.247  & \textbf{0.973} & 0.468 & 0.788 & 0.362 \\
\rowcolor{green!30} ruT5-detox & \textbf{0.738} & \textbf{0.763} & \textbf{0.807} & \textbf{0.453}  & \multicolumn{4}{c}{---} \\
\rowcolor{green!30} BART-detox & \multicolumn{4}{c|}{---}  & 0.892 & \textbf{0.624} & \textbf{0.833} & \textbf{0.458}  \\ \hline

\multicolumn{9}{c}{\it Multilingual models trained on parallel monolingual corpora} \\ \hline
mBART RU & 0.672 & 0.750 & 0.781 & 0.392 & \multicolumn{4}{c}{---} \\
mBART EN & \multicolumn{4}{c|}{---} & 0.857 & 0.599 & 0.824 & 0.418 \\
mBART EN+RU & 0.660 & \textbf{0.758} & \textbf{0.784} & 0.392  & \textbf{0.884} & 0.599 & \textbf{0.835} & \textbf{0.435} \\
M2M100+Adapter & \textbf{0.709} & 0.747 & 0.754 & \textbf{0.397} &  0.876 & 0.601 & 0.785 & 0.413 \\
mBART*+Adapter & 0.650 & \textbf{0.758} & 0.778 & 0.383  & 0.863 & \textbf{0.617} & 0.829 & \textbf{0.435}  \\
\hline \hline

\multicolumn{9}{c}{\textbf{Cross-lingual Text Detoxification Transfer} (from a language with to a language without a parallel corpus)} \\ \hline
\multicolumn{9}{c}{\it Backtranslation: monolingual model wrapped by two translations} \\ \hline
\rowcolor{green!30} ruT5-detox (FSMT) & \multicolumn{4}{c|}{---} & \textbf{0.680} & 0.458 & 0.902 & \textbf{0.324}  \\
\rowcolor{green!30} BART-detox (Yandex) & \textbf{0.601} & 0.709 & 0.832 & \textbf{0.347}  & \multicolumn{4}{c}{---} \\
mBART (Yandex) & 0.595 & 0.710 & \textbf{0.835} & 0.345 & 0.661 & \textbf{0.561} & \textbf{0.913} & 0.322 \\
\hline

\multicolumn{9}{c}{\it Translation of parallel corpus and training model on it} \\ \hline
mBART RU-Tr (Helsinki) & 0.429 & 0.773 & 0.780 & 0.257 & \multicolumn{4}{c}{---} \\
mBART EN-Tr (FSMT) & \multicolumn{4}{c|}{---} & \textbf{0.762} & 0.553 & \textbf{0.871} & \textbf{0.354} \\
\hline

\multicolumn{9}{c}{\it Multitask learning: translation of parallel corpus and adding relevant datasets } \\ \hline
mBART EN+RU-Tr & \textbf{0.552} & 0.749 & \textbf{0.783} & \textbf{0.320}  & \multicolumn{4}{c}{---} \\
mBART EN-Tr+RU &\multicolumn{4}{c|}{---} & 0.539 & 0.749 & 0.783 & 0.312  \\
\hline

\multicolumn{9}{c}{\it Adapter training: training multilingual models on monolingual corpus w/o translation} \\ \hline
M2M100+Adapter RU & \multicolumn{4}{c|}{---} & 0.422 & \textbf{0.630} & 0.779 & 0.186 \\ 
M2M100+Adapter EN & 0.340 & \textbf{0.722} & 0.675 & 0.160 & \multicolumn{4}{c}{---} \\
mBART*+Adapter RU & \multicolumn{4}{c|}{---} & \textbf{0.697} & 0.570 & \textbf{0.847} & \textbf{0.315} \\
mBART*+Adapter EN & \textbf{0.569} & 0.705 & \textbf{0.776} & \textbf{0.303} & \multicolumn{4}{c}{---} \\
\hline \hline

\multicolumn{9}{c}{\textbf{Detox\&Translation: Simultaneous Text Detoxification and Translation} } \\ 
\hline

\multicolumn{9}{c}{\it Step-by-step approach: monolingual detoxifier as a pivot + translation from/to the pivot} \\ \hline

 ruT5-detox (FSMT) & \multicolumn{4}{c|}{---} & \textbf{0.930} & 0.396 & \textbf{0.794} & \textbf{0.300} \\
\rowcolor{green!30} BART-detox (Yandex) & 0.775 & \textbf{0.694} & \textbf{0.876} & \textbf{0.467} & \multicolumn{4}{c}{---} \\
\hline

\multicolumn{9}{c}{\it End-to-end models trained on cross-lingual parallel detoxification corpus} \\ \hline
\rowcolor{green!30}{mBART (Yandex)} & \textbf{0.788} & 0.562 & 0.744 & 0.333 & \textbf{0.922} & \textbf{0.446} & \textbf{0.728} & \textbf{0.305} \\
mT5 (Yandex) & 0.782 & \textbf{0.592} & \textbf{0.790} & \textbf{0.361} & 0.897 & 0.393 & 0.558 & 0.204 \\
\bottomrule
\end{tabular}
\caption{\textbf{Automatic evaluation results}.
Numbers in \textbf{bold} indicate the best results in the sub-sections. \colorbox{green!30}{Rows in green} indicate the best models per tasks. In \textit{(brackets)}, the method of translation used for the approach is indicated. EN or RU denotes training corpus language -- original monolingual ParaDetox, while EN-Tr or RU-Tr denotes translated versions of ParaDetox. mBART* states that the version of mBART fine-tuned on paraphrasing and translation data.} 
\label{tab:cross_lingual_results}
\end{table*}

\subsection{Cross-lingual Detoxification Transfer}
From Table~\ref{tab:cross_lingual_results}, we see that backtranslation approach performed with SOTA monolingual detoxification models yields the best TST scores. This is the only approach that does not require additional model fine-tuning. However, as we can see from Table~\ref{tab:resources_approaches}, it is dependent on the constant availability of translation system which concludes in three inference steps.

Training Data Translation approach for both languages shows the J score at the level of condBERT baseline. While SIM and FL scores are the same or even higher than monolingual SOTA, the STA scores drop significantly. Some toxic parts in translated sentences can be lost while translating the toxic part of the parallel corpus. It is an advantage for the Backtranslation approach as we want to reduce toxicity only in output, while for training parallel detox corpus, we lose some of the toxicity representation. However, this approach can be used as a baseline for monolingual detoxification (examples of translation outputs in Appendix~\ref{sec:app_translation_examples}). Addition of other tasks training data to a translated ParaDetox yields improvement in the performance for the Russian language in Multitask setup. Paraphrasing samples can enrich toxicity examples that cause the increment in STA. In terms of required resources, the translation system can be used only once during training data translation, but then the fine-tuning step is present in this approach.

The adapter for the M2M100 model successfully compresses detoxification knowledge but fails to transfer it to another language. The results are completely different for additionally fine-tuned mBART. This configuration outperforms all unsupervised baselines and the Training Data Translation approach. Still, the weak point for this approach and the STA score, while not all toxicity types, can be easily transferred. However, Adapter Training is the most resource-conserving approach: it does not require additional data creation and has only one inference step. The fine-tuning procedure should be cost-efficient as we freeze the layes of the base language model and back-propagate through only adapter layers. The adapter approach can be the optimal solution for cross-lingual detoxification transfer.

Finally, according to manual evaluations in Table~\ref{tab:manual_results}, Backtranslation is the best choice if we want to transfer knowledge to the English language. However, for another low-resource language, the Adapter approach seems to be more beneficial. In the Backtranlsation approach for the Russian language, we have observed a huge loss of content. That can be a case of more toxic expressions in Russian, which are hard to translate precisely into English before detoxification. As a result, we can claim that the Adapter approach is the most efficient and precise way to transfer detoxification knowledge transfer from English to other languages.


\subsection{Detox\&Translation} At the bottom of Table~\ref{tab:cross_lingual_results}, we report experiments of baseline approaches: detoxification with monolingual detoxification SOTA, then translation into the target language. 

We can observe that our proposed approaches for this task for English perform better than the baselines. While for Russian, the results are slightly worse; our models require fewer computational resources during inference. Thus, we can claim that simultaneous style transfer with translation is possible with multilingual LM.

\section{Conclusion}
We present the first of our knowledge extensive study of cross-lingual text detoxification approaches. 
The automatic evaluation shows that the Backtranslation approach achieves the highest performance. However, this approach is bounded to the translation system availability and requires three steps during inference. The Training Data Translation approach can be a good baseline for a separate monolingual detoxification system in the target language. On the other hand, the Adapter approach requires only one inference step and performs slightly worse than Backtranslation. The adapter method showed the best manual evaluation scores when transferring from English to Russian. However, the open challenge is the capturing of the whole scope of toxicity types in the language. 

We present the first study of detoxification and translation in one step. We show that the generation of a synthetic parallel corpus where the toxic part is in one language, and the non-toxic is in another using NMT is effective for this task. Trained on such a corpus, multilingual LMs perform at the level of the backtranslation requiring fewer computations. 

All information about datasets, models, and evaluation metrics can be found online.\footnote{\href{https://github.com/dardem/text_detoxification}{https://github.com/dardem/text\_detoxification}}$^,$\footnote{\href{https://github.com/s-nlp/multilingual_detox}{https://github.com/s-nlp/multilingual\_detox}}

\section*{Acknowledgements}

We thank Elisei Stakovskii for manual evaluation of the detoxification models outputs of this paper.

\begin{table}[t!]
    \centering
    \scalebox{0.8}{
    \begin{tabular}{l|c|c|c|c}
    \toprule
         & \textbf{STA} & \textbf{SIM} & \textbf{FL} & \textbf{J} \\
        \hline
         & \multicolumn{4}{c}{English} \\
        \hline
        \rowcolor{gray!25} BART-detox (monolingual) & 0.94 & 0.96 & 1.00 & 0.90 \\
        Backtr. ruT5-detox (FSMT) & \textbf{0.78} & \textbf{0.78} & \textbf{1.00} & \textbf{0.58} \\
        mBART+Adapter RU & 0.74 & 0.70 & 0.96 & 0.42 \\
        \hline
        & \multicolumn{4}{c}{Russian} \\
        \hline
        \rowcolor{gray!25} ruT5-detox (monolingual) & 0.84 & 0.96 & 1.00 & 0.82 \\ 
        Backtr. BART-detox (Yandex) & 0.78 & 0.56 & \textbf{1.00} & 0.40 \\
        mBART+Adapter EN & \textbf{0.80} & \textbf{0.92} & 0.96 & \textbf{0.72} \\
    \bottomrule
    \end{tabular}
    }
    \caption{\textbf{Manual evaluation results}. We report the SOTA monolingual models for each language for reference and the best multilingual models (based on Backtranslation and Adapter approaches).}
    \label{tab:manual_results}
\end{table}

\section{Limitations}
One limitation of this work is the usage of only two languages for our experiments -- English and Russian. There is a great opportunity for improvement to experiment with more languages and their pairs to transfer knowledge in a cross-lingual style. 
The possibility of solving the detoxification task, requires the presence of a corpus of toxicity classification for the language. Firstly, creating a test set and building a classifier for STA evaluation is necessary. Also, having some embedding model for the language is important to calculate the SIM score for evaluation. For FL, in this work, we use classifiers. However, such classifiers can not be present in other languages. 



\section*{Ethical Considerations}

Text detoxification  has various  applications, e.g. moderating output of generative neural networks to prevent reputation losses of companies. Think of a chatbot responding rudely. Yet automatic detoxification of user content should be done with extreme care. Instead, a viable use-case is to suggest that the user rewrite a toxic comment (e.g., to save her digital reputation as the `internet remembers everything'). It is  crucial to leave the freedom to a person to express comment in the way she wants, given legal boundaries. 


\bibliography{anthology,custom}
\bibliographystyle{acl_natbib}

\newpage

\onecolumn
\appendix

\section{Examples of Detoxification Models Outputs}
\label{sec:app_examples}

\begin{table*}[h!]
    \footnotesize
    \centering
    \scalebox{0.79}{
    \begin{tabular}{c|p{7.5cm}|p{7.5cm}}
    \hline
        \textbf{Model} & \textbf{Source} & \textbf{Generated} \\ 
        \hline
        \multicolumn{3}{c}{\textbf{Monolingual Models}} \\
        \hline
        ruT5-detox & \foreignlanguage{russian}{Судя по -всему, дебилы, которые готовы отдать за этот таз 1млн. руб закончились, а бабки как то осваивать надо} \newline \tiny{\textit{Apparently, they are morons who are ready to give 1 million for this basin. the rubles are over, but we need to master the money somehow}} & \foreignlanguage{russian}{Судя по всему те люди, которые готовы отдать за этот таз 1 млн руб закончились, а деньги как то осваивать надо} \newline \tiny{\textit{Apparently, those people who are ready to pay 1 million rubles for this basin have run out, but the money needs to be used somehow}} \\ 
         & \foreignlanguage{russian}{А учиться чему что уехала за бугор бл**ь что она в беларуси не осталось и дальше не боролась оппозиция называется} \newline \tiny{\textit{And what did the b*tch go over the hill to learn, that she didn’t remain in Belarus and didn’t fight any further, the opposition is called}} & \foreignlanguage{russian}{А учиться чему? Что уехала заграницу, что она в Беларуси не осталась и дальше не боролась. Оппозиция называется} \newline \tiny{\textit{What to study? That she went abroad, that she did not stay in Belarus and did not fight further. The opposition is called}}\\ 
         & \foreignlanguage{russian}{П**деж. Бабка бы удавилась за свои кровные. Вброс расчитан на тупое быдло.} \newline \tiny{\textit{F*ck. Grandma would have strangled herself for her money. The stuffing is designed for stupid rednecks.}} & \foreignlanguage{russian}{вброс. бабка бы удавилась за свои кровные. вброс расчитан на глупых людей} \newline \tiny{\textit{bullsh*t. Grandma would have strangled herself for her money. stuffing is designed for stupid people}} \\ 
         \hline
        BART-detox & He is a f**king parasite. & He is not good. \\ 
         & This is a joke , are you all f**king retards? & This is a joke, are you all naive? \\ 
         & They are two faced f**ks. & They are two faced people. \\ \hline
        \multicolumn{3}{c}{\textbf{Backtranslation}} \\ \hline
        ruT5-detox (FSMT) & He is a f**king parasite. & He is not a good person, he is a very bad person. \\ 
         & This is scaring the s**t out of me. & This scares the hell out of me. \\ 
         & They are two faced f**ks. & They are bad people \\ \hline
        BART-detox (Yandex) & \foreignlanguage{russian}{Лучше б руководство выкинули бы на хрен!!!!} \newline \tiny{\textit{It would be better if the management were thrown out to hell!!!!}} & \foreignlanguage{russian}{Лучше бы руководство уволили!!!!} \newline \tiny{\textit{It would be better if the management was fired!!!!}} \\ 
         & \foreignlanguage{russian}{Ты бы уже п**ло язык за зубами держал} \newline \tiny{\textit{You should have kept your mouth shut, you bastard}} & \foreignlanguage{russian}{Ты бы держал рот на замке} \newline\tiny{\textit{You should keep your mouth shut}} \\ 
         & \foreignlanguage{russian}{Вот таким е**нашкам ещё права дают!!!} \newline \tiny{\textit{They still give rights to such motherf*ckers!!!}} & \foreignlanguage{russian}{Дают права таким людям! Они не милые.} \newline \tiny{\textit{Give rights to such people! They're not cute.}}\\ 
         \hline
        \multicolumn{3}{c}{\textbf{Multitask Learning}} \\ 
        \hline
        mBART RU-Tr (Helsinki) & \foreignlanguage{russian}{Г**но,оно и в Африке г**но.} \newline\tiny{\textit{It's sh*t, it's sh*t in Africa too.}} & \foreignlanguage{russian}{Это и в Африке нехорошо.} \newline \tiny{\textit{This is not good in Africa either.}} \\ 
         & \foreignlanguage{russian}{П**оры живущие за счёт народа, в биологии называются паразитами, организм живущий за счёт другова} \newline \tiny{\textit{F*gs living at the expense of the people are called parasites in biology, an organism living at the expense of others}} & \foreignlanguage{russian}{Люди живущие за счёт народа, в биологии называются паразитами, организм живущий за счет другого.} \newline \tiny{\textit{People living at the expense of the people are called parasites in biology, an organism living at the expense of another.}} \\ 
         & \foreignlanguage{russian}{А потом подобные идиоты Сапсаны под откос пускают.} \newline \tiny{\textit{And then idiots like Sapsan derail them.}} & \foreignlanguage{russian}{А потом такие люди под откос пускают, как Сапсан.} \newline \tiny{\textit{And then people like Sapsan derail it.}} \\
         \hline
        mBART EN-Tr (FSMT) & \foreignlanguage{russian}{Вот х**и вам бабам еще надо? такой прискурант озвучил!} \newline \tiny{What the f*ck do you women still need? such a price list was announced!}& \foreignlanguage{russian}{вот что вам еще надо? такой прискурант озвучил!} \newline \tiny{\textit{what else do you need? such a price list was announced!}} \\ 
         & \foreignlanguage{russian}{Политика это вообще один большой фейспалм, стадо п**оров, на**ывающих друг друга.} \newline \tiny{Politics is generally one big facepalm, a herd of f*ggots f*cking each other.}& \foreignlanguage{russian}{Политика - это вообще один большой фейспалм, где люди разговаривают друг с другом.} \newline \tiny{\textit{Politics is generally one big facepalm where people talk to each other.}} \\ 
         & \foreignlanguage{russian}{Как можно было такую уродку выдвигать в депутаты?} \newline \tiny{\textit{How could such a freak be nominated as a deputy?}} & \foreignlanguage{russian}{Как можно было её выдвигать в депутаты?} \newline \tiny{\textit{How could she be nominated as a deputy?}}\\ 
        \hline
        \multicolumn{3}{c}{\textbf{Adapter Training}}\\
        \hline
        mBART*+Adapter RU & \foreignlanguage{russian}{Вот х**и вам бабам еще надо? такой прискурант озвучил!} \newline \tiny{\textit{What the f*ck do you women still need? such a price list was announced!}} & \foreignlanguage{russian}{Вот что вам женщинам ещё надо? Такой прейскурант озвучил!} \newline \tiny{\textit{what else do you need? such a price list was announced!}}\\ 
         & \foreignlanguage{russian}{вот подлец ,разыграть меня хочет ,старьё мне подсовывает} \newline \tiny{\textit{here's a scoundrel, he wants to play a trick on me, he gives me old stuff}} & \foreignlanguage{russian}{Разыграть меня хочет, старьё мне подсовывает.} \newline \tiny{\textit{He wants to play a trick on me, he gives me old things.}}\\ 
         & \foreignlanguage{russian}{женщина изменяет ибо она б**дь} \newline \tiny{\textit{a woman cheats because she's a b**ch}} & \foreignlanguage{russian}{Женщина изменяет ибо она неверная} \newline \tiny{\textit{A woman cheats because she is unfaithful}} \\ \hline
        mBART*+Adapter EN & because israeli rabbis never say f**ked up s**t. & Because Israeli rabbis don't say bad things \\ 
         & cretins like this rarely care about facts. & People like this rarely care about facts. \\ 
         & so , 'cctv shows' crimea parliament explosion with a shitty picture of fuck knows what with a bit of smoke in it . & so, 'cctv shows' crimea parliament explosion with a bad picture of God knows what with a bit of smoke in it. \\
         \hline
        \multicolumn{3}{c}{\textbf{Detoxification with Translation}} \\ \hline
        ruT5-detox (FSMT) & learn how to write in Russian, you f**king f**got. & \foreignlanguage{russian}{Научитесь писать по-русски.} \newline \tiny{Learn to write in Russian.} \\ 
         & better get the hell out of here you don't belong here & \foreignlanguage{russian}{Уходи, здесь нет места для тебя.} \newline \tiny{\textit{Go away, there's no place for you here.}}\\ 
         & Do you like boys like that, motherf**ker? & \foreignlanguage{russian}{Тебе нравятся эти мальчики?} \newline \tiny{\textit{Do you like these boys?}} \\ 
         \hline
        BART-detox (Yandex) & actually no , i hate looking for s**t. & \foreignlanguage{russian}{На самом деле нет, я ненавижу это искать.} \newline \tiny{\textit{Actually no, I hate looking it up.}}\\ 
         & calm the f**k down, cnn. & \foreignlanguage{russian}{успокойся, cnn.com.}\newline \tiny{\textit{Rest easy, cnn.com.}} \\ 
         & get this ldiot out of there . & \foreignlanguage{russian}{Вытащи этого парня оттуда.} \newline \tiny{\textit{Get this guy out of there.}} \\ 
         \hline
        mBART (Yandex) & \foreignlanguage{russian}{не надо опускаться до уровня шута безмозглого} \newline \tiny{\textit{no need to stoop to the level of a brainless buffoon}} & It is not necessary to descend to the lower level \\ 
         & \foreignlanguage{russian}{вот твари. и как они после этого заснуть смогут.} \newline \tiny{\textit{here are the creatures. and how will they be able to sleep after that?}} & And how will they be able to fall asleep after that \\ 
         & \foreignlanguage{russian}{ну и е**нутая же ты б**ть} \newline \tiny{\textit{what a f*cking nut you are}} & Well, you're a weird person. \\ 
         \hline
        mT5 (Yandex) & \foreignlanguage{russian}{тьфу б**дь сопля, иди работай чмо!} \newline \tiny{\textit{ugh f*cking snot, go to work schmuck!}} & Leave me alone, go to work. I don't like you \\ 
         & \foreignlanguage{russian}{твари уже не знают как ещё над людьми поиздеваться:(} \newline \tiny{\textit{the creatures no longer know how else to make fun of people :(}}& They don't know how to make fun of people anymore \\ 
         & \foreignlanguage{russian}{х**ня шо больше показать нечего} \newline \tiny{\textit{crap, nothing more to show}} & There's nothing else to show. \\
        \hline
    \end{tabular}}
    \caption{Examples of models' output: baselines and our proposed approaches.}
    \label{tab:app_outputs}
\end{table*}

\clearpage

\section{Examples of Translation Outputs}
\label{sec:app_translation_examples}

\begin{table}[h!]
\centering
\footnotesize
\begin{tabular}{p{3cm}|p{5cm}|p{5cm}|p{1cm}}
\toprule
\textbf{Translator} & \textbf{Source} & \textbf{Translated} & \textbf{Quality} \\
\hline
FSMT \cite{DBLP:conf/wmt/NgYBOAE19} & what the f**k is this comment ? &  \foreignlanguage{russian}{Что это за х**ня?} & \redcross \\
FSMT & the fact we are doing this now and it started when i was number number is fucking insane ! & \foreignlanguage{russian}{тот факт, что мы делаем это сейчас, и это началось, когда я был номером, это гребано!} & \redcross \\
FSMT & the germans today are the same blind brainless followtheleader sheep they were in and throughout their history. & \foreignlanguage{russian}{сегодняшние немцы такие же слепые безмозглые овчарки, какими они были в своей истории.} & \greencheck \\
FSMT & the guy was f**king shot ! & \foreignlanguage{russian}{парень гребаный стрелял!} & \redcross \\
FSMT & maybe now no one will be rammed in the *ss . & \foreignlanguage{russian}{Может быть, теперь никто не будет ругаться в з**ницу.} & \redcross \\
\midrule
OPUS-MT \cite{DBLP:conf/eamt/TiedemannT20} & what the f**k is this comment ? &  \foreignlanguage{russian}{Что это за х**ня?} & \redcross \\
OPUS-MT c& the fact we are doing this now and it started when i was number number is f**king insane !
 & \foreignlanguage{russian}{То, что мы делаем сейчас и это началось когда я был номером - сумасшествие!} & \redcross \\
OPUS-MT & the germans today are the same blind brainless followtheleader sheep they were inand throughout their history . & \foreignlanguage{russian}{Сегодня немцы - те же слепые безмозглые овцы, что и во всей их истории.
} & \greencheck \\
OPUS-MT & the guy was f**king shot ! & \foreignlanguage{russian}{Парня застрелили!} & \redcross \\
OPUS-MT & maybe now no one will be rammed in the a** . & \foreignlanguage{russian}{Может быть, теперь никто не будет запихнут в ж**у.} & \redcross \\
OPUS-MT & it s actually a great idea if you want to keep nuts out and get s**t done . & \foreignlanguage{russian}{Это действительно отличная идея, если ты хочешь держаться подальше и заниматься д*рьмом.} & \redcross \\
\midrule
Yandex.Translate & what the f**k is this comment ? & \foreignlanguage{russian}{что, черт возьми, это за комментарий?} & \greencheck \\
Yandex.Translate  & the fact we are doing this now and it started when i was number number is f**king insane ! & \foreignlanguage{russian}{тот факт, что мы делаем это сейчас, и это началось, когда я был номером номер, чертовски безумен!} & \greencheck \\
Yandex.Translate & the germans today are the same blind brainless followtheleader sheep they were inand throughout their history . & \foreignlanguage{russian}{немцы сегодня - такие же слепые безмозглые овцы, следующие за лидером, какими они были на протяжении всей своей истории.} & \greencheck \\
Yandex.Translate & the guy was f**king shot ! & \foreignlanguage{russian}{этого парня, б**дь, застрелили!} & \greencheck \\
Yandex.Translate & maybe now no one will be rammed in the a** . & \foreignlanguage{russian}{может быть, теперь никого не будут таранить в з*дницу.}  & \redcross \\
Yandex.Translate & it s actually a great idea if you want to keep nuts out and get s**t done . & \foreignlanguage{russian}{на самом деле это отличная идея, если вы хотите не сходить с ума и довести дело до конца.} & \greencheck \\
\midrule
Google.Translate & what the f**k is this comment ? & \foreignlanguage{russian}{что за бред этот комментарий?} & \greencheck \\
Google.Translate  & the fact we are doing this now and it started when i was number number is f**king insane ! & \foreignlanguage{russian}{тот факт, что мы делаем это сейчас, и это началось, когда я был номером номер, чертовски безумен!} & \greencheck \\
Google.Translate & the germans today are the same blind brainless followtheleader sheep they were inand throughout their history . & \foreignlanguage{russian}{нынешние немцы - такие же слепые безмозглые овцы, следующие за вожаками, которыми они были на протяжении всей своей истории.} & \greencheck \\
Google.Translate & the guy was f**king shot ! & \foreignlanguage{russian}{парень был чертовски застрелен!} & \greencheck \\
Google.Translate & maybe now no one will be rammed in the a** . & \foreignlanguage{russian}{может теперь никто не будет таранить под з*д.}  & \redcross \\
Google.Translate & it s actually a great idea if you want to keep nuts out and get s**t done . & \foreignlanguage{russian}{на самом деле это отличная идея, если вы хотите держаться подальше от орехов и делать д*рьмо.} & \greencheck \\
\bottomrule
\end{tabular}
\caption{Examples of translations from English to Russian.}
\end{table}

\begin{table}[h!]
\footnotesize
\begin{tabular}{p{3cm}|p{5cm}|p{5cm}|p{1cm}}
\toprule
\textbf{Translator} & \textbf{Source} & \textbf{Translated} & \textbf{Quality} \\
\hline
FSMT \cite{DBLP:conf/wmt/NgYBOAE19} & \foreignlanguage{russian}{бл**ь, ты хоть себя слышишь?)} \newline \foreignlanguage{russian}{ты говоришь что я экстрасенс, а потом говоришь, что нет} & Do you even hear yourself?) \newline You say I'm a psychic, and then you say no. & \redcross \\
FSMT \cite{DBLP:conf/wmt/NgYBOAE19} &\foreignlanguage{russian}{лично я хочу чтоб мр*зи сели на пожизненое} & Personally, I want them to sit down for life. & \redcross \\
FSMT \cite{DBLP:conf/wmt/NgYBOAE19} & \foreignlanguage{russian}{тварь,трус! ничего человеческого не осталось} & Creature, c*ward! There is nothing human left. & \redcross \\
FSMT \cite{DBLP:conf/wmt/NgYBOAE19} & \foreignlanguage{russian}{От этого пострадают только всякие усть-переп**дюйск-телекомы с 3.5 сотрудниками} & Only those with 3.5 employees will be affected. & \greencheck \\
FSMT \cite{DBLP:conf/wmt/NgYBOAE19} & \foreignlanguage{russian}{иди н**ер, верните иваныча, черти!} & Go n**her, bring back Ivanich, devils! & \greencheck \\
\midrule
OPUS-MT \cite{DBLP:conf/eamt/TiedemannT20} & \foreignlanguage{russian}{бл**ь, ты хоть себя слышишь?) ты говоришь что я экстрасенс, а потом говоришь, что нет} & Can you f**king hear yourself?) You say I'm a psychic, and then you tell me no. & \greencheck \\
OPUS-MT \cite{DBLP:conf/eamt/TiedemannT20} & \foreignlanguage{russian}{лично я хочу чтоб мр*зи сели на пожизненое} & Personally, I want the b*stards to sit down for life. & \greencheck \\
OPUS-MT \cite{DBLP:conf/eamt/TiedemannT20} & \foreignlanguage{russian}{тварь,трус! ничего человеческого не осталось} & You son of a b**ch! There's nothing human left. & \greencheck \\
OPUS-MT \cite{DBLP:conf/eamt/TiedemannT20} & \foreignlanguage{russian}{От этого пострадают только всякие усть-переп**дюйск-телекомы с 3.5 сотрудниками} & This will only cause damage to any of the three-way telecoms with 3.5 employees. & \redcross \\
OPUS-MT \cite{DBLP:conf/eamt/TiedemannT20} & \foreignlanguage{russian}{эти бл**и совсем о**ели тв*ри конченые} & These f**king things are so f**ked up.  & \redcross \\
OPUS-MT \cite{DBLP:conf/eamt/TiedemannT20} & \foreignlanguage{russian}{иди н**ер, верните иваныча, черти!} & Go f**k yourself, get the Ivanich back! & \redcross \\
\midrule
Yandex.Translate & \foreignlanguage{russian}{бл**ь, ты хоть себя слышишь?) ты говоришь что я экстрасенс, а потом говоришь, что нет} & Can you f**king hear yourself?) You say I'm a psychic, and then you tell me no. & \greencheck \\
Yandex.Translate  & \foreignlanguage{russian}{лично я хочу чтоб мр*зи сели на пожизненое} & Personally, I want the sc*m to go to prison for life. & \greencheck \\
Yandex.Translate & \foreignlanguage{russian}{тварь,трус! ничего человеческого не осталось} & You coward! There's nothing human left. & \greencheck \\
Yandex.Translate & \foreignlanguage{russian}{От этого пострадают только всякие усть-переп**дюйск-телекомы с 3.5 сотрудниками} & Only Ust-perep**dyuisk telecoms with 3.5 employees will suffer from this & \greencheck \\
Yandex.Translate & \foreignlanguage{russian}{эти бляди совсем о**ели твари конченые} & these whores are completely f**ked up creatures are finished  & \redcross \\
Yandex.Translate & \foreignlanguage{russian}{иди н**ер, верните иваныча, черти!} & go to hell, bring Ivanovich back, damn it! & \greencheck \\
\midrule
Google.Translate & \foreignlanguage{russian}{бл**ь, ты хоть себя слышишь?) ты говоришь что я экстрасенс, а потом говоришь, что нет} & f**k, can you even hear yourself?) you say that I'm a psychic, and then you say that I'm not & \greencheck \\
Google.Translate & \foreignlanguage{russian}{лично я хочу чтоб мр*зи сели на пожизненое} & I personally want the sc*m to sit on a life sentence & \greencheck \\
Google.Translate & \foreignlanguage{russian}{тварь,трус! ничего человеческого не осталось} & creature, c*ward! nothing human left & \greencheck \\
Google.Translate & \foreignlanguage{russian}{От этого пострадают только всякие усть-переп**дюйск-телекомы с 3.5 сотрудниками} & Only all sorts of Ust-Perep**duysk-Telecoms with 3.5 employees will suffer from this & \greencheck \\
Google.Translate & \foreignlanguage{russian}{эти бл**и совсем охуели тв*ри конченые} & these whores are completely f**ked up by the finished creatures  & \redcross \\
Google.Translate & \foreignlanguage{russian}{иди н**ер, верните иваныча, черти!} & go to hell, bring Ivanovich back, d*mn it!& \greencheck \\
\bottomrule
\end{tabular}
\caption{Examples of translations from Russian to English.}
\end{table}

\clearpage 



\section{Human vs Automatic Evaluation Correlations for Old and New Setups}
\label{sec:app_correlations}

The detailed correlation results of new and old automatic metrics for the Russian language: (i)~based on system score (Table~\ref{tab:auto_vs_manual_scores_ru}); (ii)~based on system ranking (Table~\ref{tab:auto_vs_manual_ranking_ru}). 

In the first approach, we concatenate all the scores of all systems for corresponding metrics in one vector and calculate Spearman's correlation between such vectors for human and automatic evaluation. For the second approach, we rank the systems based on the corresponding metric, get the vector of the systems' places in the leaderboard, and calculate Spearman's correlation between such vectors for human and automatic evaluation. We can observe improvements in correlations for both setups with newly presented metrics.

\begin{table}[h!]
\centering
\footnotesize
\begin{tabular}{crrrr}
\toprule
Metric & \multicolumn{1}{c}{STA$_a^{old}$} & \multicolumn{1}{c}{SIM$_a^{old}$} & \multicolumn{1}{c}{FL$_a^{old}$} & \multicolumn{1}{c}{J$_a^{old}$} \\
\midrule
STA$_{m}$ & 0.472 & \textbf{-0.324} & -0.121 & 0.120 \\
SIM$_{m}$ & \textbf{-0.062} & 0.124 & 0.084 & -0.026 \\
FL$_{m}$ & 0.018 & \textbf{-0.087} & -0.011 & -0.132 \\
J$_{m}$ & 0.271 & \textbf{-0.138} & -0.031 & 0.106 \\
\midrule
Metric & \multicolumn{1}{c}{STA$_a$} & \multicolumn{1}{c}{SIM$_a$} & \multicolumn{1}{c}{FL$_a$} & \multicolumn{1}{c}{J$_a$} \\
\midrule
STA$_{m}$ & \textbf{0.598} & -0.071 & \textbf{0.130} & \textbf{0.516} \\
SIM$_{m}$ & -0.012 & \textbf{0.244} & \textbf{0.217} & \textbf{0.176} \\
FL$_{m}$ & \textbf{0.107} & 0.054 & \textbf{0.354} & \textbf{0.229} \\
J$_{m}$ & \textbf{0.370} & 0.096 & \textbf{0.259} & \textbf{0.482} \\
\bottomrule
\end{tabular}
\caption{Spearman's correlation coefficient between automatic VS manual metrics based on systems scores for \textbf{Russian} language. All numbers denote the statistically significant correlation ($p$-value $\leq 0.05$).}
\label{tab:auto_vs_manual_scores_ru}
\end{table}

\begin{table}[h!]
\centering
\footnotesize
\begin{tabular}{crrrr}
\toprule
Metric & \multicolumn{1}{c}{STA$_a^{old}$} & \multicolumn{1}{c}{SIM$_a^{old}$} & \multicolumn{1}{c}{FL$_a^{old}$} & \multicolumn{1}{c}{J$_a^{old}$} \\
\midrule
STA$_{m}$ & 0.235 & \textbf{-0.657} & -0.200 & 0.138 \\
SIM$_{m}$ & 0.130 & 0.015 & 0.240 & 0.248 \\
FL$_{m}$ & -0.024 & -0.284 & 0.024 & 0.002 \\
J$_{m}$ & 0.169 & -0.116 & 0.204 & 0.231 \\
\midrule
Metric & \multicolumn{1}{c}{STA$_a$} & \multicolumn{1}{c}{SIM$_a$} & \multicolumn{1}{c}{FL$_a$} & \multicolumn{1}{c}{J$_a$} \\
\midrule
STA$_{m}$ & \textbf{0.811} & -0.231 & \textbf{0.600} & \textbf{0.692} \\
SIM$_{m}$ & \textbf{0.240} & \textbf{0.732} & \textbf{0.349} & \textbf{0.648} \\
FL$_{m}$ & \textbf{0.292} & \textbf{0.305} & \textbf{0.868} & \textbf{0.613} \\
J$_{m}$ & \textbf{0.433} & \textbf{0.565} & \textbf{0.534} & \textbf{0.802} \\
\bottomrule
\end{tabular}
\caption{Spearman's correlation coefficient between automatic VS manual metrics based on system ranking for \textbf{Russian} language. All numbers denote the statistically significant correlation ($p$-value $\leq 0.05$)}
\label{tab:auto_vs_manual_ranking_ru}
\end{table}

\clearpage

\section{Comparison of Translation Methods}
\label{sec:app_translation}
Here, we provide a thorough comparison of all mentioned translation methods for presented approaches: (i) Cross-lingual Detoxification Transfer (Table~\ref{tab:cross_lingual_results_app}); (ii) Detox\&Translation (Table~\ref{tab:detox_translation_results_app}). Additionally, we provide the experiments for \textit{multilingual} setup (where the detoxification models are trained on datasets in both languages simultaneously) for \textit{Training Data Translation} approach in Table~\ref{tab:multilingual_results_app}.

\begin{table*}[h!]
\footnotesize
\centering
\begin{tabular}{p{3.65cm}|c|c|c|c|c|c|c|c}
\toprule
 & \textbf{STA} & \textbf{SIM} & \textbf{FL} & \textbf{J} & \textbf{STA} & \textbf{SIM} & \textbf{FL} & \textbf{J} \\
\hline
 & \multicolumn{4}{c|}{\textbf{Russian}} &
  \multicolumn{4}{c}{\textbf{English}} \\
\hline
 & \multicolumn{8}{c}{\textbf{Cross-lingual Detoxification Transfer}} \\
\hline
 & \multicolumn{8}{c}{\it Backtranslation} \\
\hline
\rowcolor{green!30} ruT5-detox (FSMT) & \multicolumn{4}{c|}{---} & \textbf{0.680} & 0.458 & \textbf{0.902} & \textbf{0.324} \\
ruT5-detox (Google) & \multicolumn{4}{c|}{---} & 0.643 & 0.565 & 0.884 & 0.311 \\
ruT5-detox (Yandex) & \multicolumn{4}{c|}{---} & 0.627 & \textbf{0.579} & 0.896 & 0.316 \\
ruT5-detox (Helsinki) & \multicolumn{4}{c|}{---} & 0.631 & 0.544 & 0.892 & 0.297 \\
\hline
BART-detox (FSMT) & 0.547 & 0.628 & 0.772 & 0.258 & \multicolumn{4}{c}{---} \\
BART-detox (Google) & 0.578 & \textbf{0.721} & 0.815 & 0.333 & \multicolumn{4}{c}{---} \\
\rowcolor{green!30} BART-detox (Yandex) & 0.601 & 0.709 & \textbf{0.832} & \textbf{0.347} & \multicolumn{4}{c}{---} \\
BART-detox (Helsinki) & \textbf{0.607} & 0.591 & 0.776 & 0.277 & \multicolumn{4}{c}{---} \\
\hline
mBART (FSMT) & 0.545 & 0.629 & 0.781 & 0.263 & \textbf{0.706} & 0.460 & 0.844 & 0.269 \\
mBART (Helsinki) & \textbf{0.599} & 0.598 & 0.774 & 0.276 & 0.671 & 0.503 & 0.859 & 0.285 \\
\rowcolor{green!30} mBART (Yandex) & 0.595 & 0.710 & \textbf{0.835} & \textbf{0.345} & 0.661 & \textbf{0.561} & \textbf{0.913} & \textbf{0.322} \\
mBART (Google) & 0.566 & \textbf{0.722} & 0.808 & 0.325 & 0.668 & 0.547 & 0.887 & 0.312 \\
\hline
 & \multicolumn{8}{c}{\it Training Data Translation} \\
\hline
mBART RU-Tr (FSMT) & \textbf{0.432} & 0.758 & \textbf{0.781} & 0.253 & \multicolumn{4}{c}{---} \\
mBART RU-Tr (Yandex) & 0.384 & \textbf{0.773} & 0.780 & 0.228 & \multicolumn{4}{c}{---} \\
\rowcolor{green!30} mBART RU-Tr (Helsinki) & 0.429 & \textbf{0.773} & 0.780 & \textbf{0.257} & \multicolumn{4}{c}{---} \\
\hline
\rowcolor{green!30} mBART EN-Tr (FSMT) & \multicolumn{4}{c|}{---} & \textbf{0.762} & 0.553 & \textbf{0.871} & \textbf{0.354} \\
mBART EN-Tr (Yandex) & \multicolumn{4}{c|}{---} & 0.648 & \textbf{0.623} & 0.838 & 0.320 \\
mBART EN-Tr (Helsinki) & \multicolumn{4}{c|}{---} & 0.646 & 0.618 & 0.858 & 0.319  \\ 
\bottomrule
\end{tabular}
\caption{Evaluation of TST models.
Numbers in \textbf{bold} indicate the best results by each parameter inside of the subsections. \colorbox{green!30}{Rows in green} indicate the best models chosen for the main results comparison. EN-Tr or RU-Tr denote translated versions of ParaDetox.} 
\label{tab:cross_lingual_results_app}
\end{table*}

\begin{table*}[h!]
\footnotesize
\centering
\begin{tabular}{p{3.65cm}|c|c|c|c|c|c|c|c}
\toprule
 & \textbf{STA} & \textbf{SIM} & \textbf{FL} & \textbf{J} & \textbf{STA} & \textbf{SIM} & \textbf{FL} & \textbf{J} \\
\hline
 & \multicolumn{4}{c|}{\textbf{Russian}} &
  \multicolumn{4}{c}{\textbf{English}} \\
\hline
 & \multicolumn{8}{c}{\textbf{Detox\&Translation}} \\
\hline
 & \multicolumn{8}{c}{\it Detoxification with Translation} \\
 \hline
 ruT5-detox (Yandex) & \multicolumn{4}{c|}{---} & 0.834 & \textbf{0.494} & 0.705 & 0.297 \\
 ruT5-detox (Google) & \multicolumn{4}{c|}{---} & 0.829 & 0.490 & 0.686 & 0.284 \\
\rowcolor{green!30} ruT5-detox (FSMT) & \multicolumn{4}{c|}{---} & \textbf{0.930} & 0.396 & \textbf{0.794} & \textbf{0.300} \\
 ruT5-detox (Helsinki) & \multicolumn{4}{c|}{---} & 0.811 & 0.442 & 0.770 & 0.279 \\
 \hline
\rowcolor{green!30} BART-detox (Yandex) & \textbf{0.774} & \textbf{0.699} & \textbf{0.876} & \textbf{0.470} & \multicolumn{4}{c}{---} \\
 BART-detox (Google) & 0.773 & 0.680 & 0.845 & 0.440 & \multicolumn{4}{c}{---} \\
 BART-detox (FSMT) & 0.674 & 0.490 & 0.802 & 0.266 & \multicolumn{4}{c}{---} \\
 BART-detox (Helsinki) & 0.674 & 0.614 & 0.802 & 0.325 & \multicolumn{4}{c}{---} \\
\hline
 & \multicolumn{8}{c}{\it Cross-lingual Training Data} \\
\hline
\rowcolor{green!30} mBART (Yandex) & \textbf{0.788} & \textbf{0.562} & \textbf{0.744} & \textbf{0.333} & \textbf{0.922} & \textbf{0.446} & \textbf{0.728} & \textbf{0.305} \\
mBART (Google) & 0.749 & 0.516 & 0.727 & 0.277 & 0.894 & 0.365 & 0.703 & 0.230 \\
\hline
mT5-base (Yandex) & \textbf{0.773} & \textbf{0.569} & \textbf{0.721} & \textbf{0.315} & \textbf{0.880} & \textbf{0.414} & \textbf{0.655} & \textbf{0.250} \\
mT5-base (Google) & 0.765 & 0.473 & 0.602 & 0.218 & 0.861 & 0.343 & 0.573 & 0.173 \\
\hline
\rowcolor{green!30} mT5-large (Yandex) & \textbf{0.782} & \textbf{0.592} & \textbf{0.790} & \textbf{0.361} & \textbf{0.897} & 0.393 & 0.558 & 0.204 \\
mT5-large (Google) & 0.745 & 0.536 & 0.708 & 0.280 & 0.846 & \textbf{0.410} & \textbf{0.713} & \textbf{0.250} \\
\bottomrule
\end{tabular}
\caption{Evaluation of TST models.
Numbers in \textbf{bold} indicate the best results by each parameter inside the subsections. \colorbox{green!30}{Rows in green} indicate the best models to compare the main results.} 
\label{tab:detox_translation_results_app}
\end{table*}

\begin{table*}[t]
\footnotesize
\centering
\begin{tabular}{p{3.65cm}|c|c|c|c|c|c|c|c}
\toprule
 & \textbf{STA} & \textbf{SIM} & \textbf{FL} & \textbf{J} & \textbf{STA} & \textbf{SIM} & \textbf{FL} & \textbf{J} \\
\hline
 & \multicolumn{4}{c|}{\textbf{Russian}} &
  \multicolumn{4}{c}{\textbf{English}} \\
\hline
 & \multicolumn{8}{c}{\textbf{Multilingual Detoxification}} \\
\hline
 & \multicolumn{8}{c}{\it Training Data Translation} \\
\hline
mBART EN+RU-Tr \newline (FSMT) & \textbf{0.490} & 0.734 & \textbf{0.788} & \textbf{0.278} & 0.863 & 0.633 & 0.838 & 0.450 \\
mBART EN+RU-Tr \newline (Yandex) & 0.410 & \textbf{0.771} & 0.786 & 0.249 & 0.852 & 0.636 & 0.826 & 0.440 \\
mBART EN+RU-Tr \newline (Helsinki) & 0.458 & \textbf{0.771} & 0.784 & 0.276 & 0.881 & 0.550 & 0.739 & 0.360 \\
\hline
mBART EN-Tr+RU \newline (FSMT) & 0.613 & 0.775 & 0.781 & 0.370 & 0.692 & 0.583 & 0.861 & 0.327 \\
mBART EN-Tr+RU \newline (Yandex) & 0.453 & 0.769 & 0.784 & 0.272 & 0.768 & \textbf{0.593} & 0.857 & 0.376 \\
mBART EN-Tr+RU \newline (Helsinki) & 0.584 & 0.780 & 0.782 & 0.356 & \textbf{0.792} & 0.583 & \textbf{0.870} & \textbf{0.386} \\
\bottomrule
\end{tabular}
\caption{Evaluation of TST models.
Numbers in \textbf{bold} indicate the best results by each parameter inside the subsections. EN-Tr or RU-Tr denote translated versions of ParaDetox.} 
\label{tab:multilingual_results_app}
\end{table*}

\clearpage 
\section{Manual Evaluation Instructions}
\label{sec:app_manual_instructions}
Here, we present the explanation of labels that annotators had to assign for each of the three evaluation parameters. We adapt the manual annotation process described in \cite{logacheva-etal-2022-study}:
\paragraph{Toxicity (STA$_m$)} \textit{Is this text offensive?}
\begin{itemize}
    \item \textbf{non-toxic} (1) --- the sentence does not contain any aggression or offence. However, we allow covert aggression and sarcasm. 
    \item \textbf{toxic} (0) --- the sentence contains open aggression and/or swear words (this also applies to meaningless sentences).
\end{itemize}

\paragraph{Content (SIM$_m$)} \textit{Does these sentences mean the same?}
\begin{itemize}
    \item \textbf{matching} (1) --- the output sentence fully preserves the content of the input sentence. Here, we allow some change of sense which is inevitable during detoxification (e.g., replacement with overly general synonyms: \textit{idiot} becomes \textit{person} or \textit{individual}). It should also be noted that content and toxicity dimensions are independent, so if the output sentence is toxic, it can still be good in terms of content. 
    \item \textbf{different} (0) --- the sense of the transferred sentence differs from the input. Here, the sense should not be confused with the word overlap. The sentence is different from its original version if its main intent has changed (cf. \textit{I want to go out} and \textit{I want to sleep}). The partial loss or change of sense is also considered a mismatch (cf. \textit{I want to eat and sleep} and \textit{I want to eat}). Finally, when the transferred sentence is senseless, it should also be considered \textit{different}.
\end{itemize}

\paragraph{Fluency (FL$_m$)} \textit{Is this text correct?}
\begin{itemize}
    \item \textbf{fluent} (1) --- sentences with no mistakes, except punctuation and capitalization errors.
    \item \textbf{partially fluent} (0.5) --- sentences with orthographic and grammatical mistakes, non-standard spellings. However, the sentence should be fully intelligible.
    \item \textbf{non-fluent} (0) --- sentences which are difficult or impossible to understand.
\end{itemize}

However, since all the input sentences are user-generated, they are not guaranteed to be fluent in this scale. People often make mistakes and typos and use non-standard spelling variants. We cannot require that a detoxification model fixes them. Therefore, we consider the output of a model fluent if the model did not make it less fluent than the original sentence. Thus, we evaluate both the input and the output sentences and define the final fluency score as \textbf{fluent} (1) if the fluency score of the output is greater or equal to that of the input, and \textbf{non-fluent} (0) otherwise.

\end{document}